\documentclass[letterpaper, 10 pt, conference]{ieeeconf_1.6}  
\IEEEoverridecommandlockouts

\pdfoutput=1
\usepackage[T1]{fontenc}
\usepackage[english]{babel}
\usepackage[pdftex]{graphicx}   
\usepackage{epsfig}             
\usepackage{amsmath}            
\usepackage{amssymb}            
\usepackage{mathtools}
\usepackage{multicol}
\usepackage{hyperref}
\usepackage{xcolor}
\usepackage{siunitx}
\usepackage{lipsum}
\usepackage{scrextend}
\usepackage{algorithm}
\usepackage{algpseudocode}

\usepackage{bm}

\usepackage{amsthm}

\theoremstyle{plain}

\theoremstyle{definition}

\theoremstyle{remark}

\newcommand{\norm}[1]{\lVert #1 \rVert}

\newcommand{\skewsym}[1]{{\lfloor #1 \rfloor}_{\times}}

\newcommand{\Rot}[2]{\prescript{#1}{#2}{\mathbf{R}}}
\newcommand{\quat}[2]{\prescript{#1}{#2}{\mathbf{q}}}
\newcommand{\Vector}[3]{\prescript{#1}{}{\bm{#2}}_{#3}}

\newcommand{\hatRot}[2]{\prescript{#1}{#2}{\hat{\mathbf{R}}}}
\newcommand{\hatVector}[3]{\prescript{#1}{}{\hat{\bm{#2}}}_{#3}}

\newcommand{\frameofref}[1]{\{$#1$\}}

\newcommand{\tagpos}{\Vector{G}{p}{U}}

\newcommand{\anchpos}{\Vector{G}{p}{A}}
\newcommand{\hatanchpos}{\hatVector{G}{p}{A}}

\newcommand{\figref}[1]{Fig.~\ref{fig:#1}}

\newcommand{\equref}[1]{Equ.~(\ref{eq:#1})}

\newcommand{\algoref}[1]{Alg.~\ref{alg:#1}}

\usepackage{acro}
\DeclareAcronym{mav} {
    short   = MAV,
    long    = Micro Aerial Vehicle,
}
\DeclareAcronym{uav} {
    short   = UAV,
    long    = Unmanned Aerial Vehicle,
}
\DeclareAcronym{imu}{
    short   = IMU,
    long    = Inertial Measurement Unit,
}
\DeclareAcronym{lidar}{
    short   = LIDAR,
    long    = Laser Imaging Detection and Ranging,
}
\DeclareAcronym{vins}{
    short   = VINS,
    long    = Visual-Inertial Navigation System,
}
\DeclareAcronym{bins}{
    short   = BINS,
    long    = biased inertial navigation system,
}
\DeclareAcronym{vinseval}{
    short   = VINSEval,
    long    = \acs{vins} Evaluation Framework,
}
\DeclareAcronym{mcsf}{
    short   = MCSF,
    long    = Multi Copter Simulation Framework,
}
\DeclareAcronym{vio}{
    short   = VIO,
    long    = Visual-Inertial Odometry,
}
\DeclareAcronym{slam}{
    short   = SLAM,
    long    = Simultaneous Localization and Mapping,
}
\DeclareAcronym{vislam}{
    short   = VI-SLAM,
    long    = Visual-Inertial \acs{slam},
}
\DeclareAcronym{ros}{
    short   = ROS,
    long    = Robot Operating System,
}
\DeclareAcronym{nees}{
    short   = NEES,
    long    = Normalized Estimation Error Squared,
}

\DeclareAcronym{anees}{
    short   = A\acs{nees},
    long    = Average \ac{nees},
}
\DeclareAcronym{rmse}{
    short   = RMSE,
    long    = Root Mean Square Error,
}
\DeclareAcronym{ate}{
    short   = ATE,
    long    = Absolute Trajectory Error,
}
\DeclareAcronym{rte}{
    short   = RTE,
    long    = Relative Trajectory Error,
}
\DeclareAcronym{bp}{
    short   = BP,
    long    = Breaking Point,
}
\DeclareAcronym{mems}{
    short   = MEMS,
    long    = Micro Electro Mechanical Systems,
}
\DeclareAcronym{mse}{
    short   = MSE,
    long    = Mean Squared Error Matrix,
}
\DeclareAcronym{eqf}{
    short   = EqF,
    long    = Equivariant Filter,
}
\DeclareAcronym{kf}{
    short   = KF,
    long    = Kalman Filter,
}
\DeclareAcronym{ekf}{
    short   = EKF,
    long    = Extended Kalman Filter,
}
\DeclareAcronym{mekf}{
    short   = MEKF,
    long    = Multiplicative \acs{ekf},
}
\DeclareAcronym{ukf}{
    short   = UKF,
    long    = Unscented Kalman Filter,
}
\DeclareAcronym{iekf}{
    short   = IEKF,
    long    = Invariant Extended Kalman Filter,
}
\DeclareAcronym{eskf}{
    short   = ESKF,
    long    = Error State Kalman Filter,
}
\DeclareAcronym{msckf}{
    short   = MSCKF,
    long    = Multi-State Constraint Kalman Filter,
}
\DeclareAcronym{uwb} {
    short   = UWB,
    long    = Ultra Wide Band,
}
\DeclareAcronym{gdop} {
    short   = GDOP,
    long    = Geometric Dilution of Precision,
}
\DeclareAcronym{fim} {
    short   = FIM,
    long    = Fisher Information Matrix,
}
\DeclareAcronym{gnss} {
    short   = GNSS,
    long    = Global Navigation Satellite System,
}
\DeclareAcronym{lm} {
    short   = LM,
    long    = Levenberg-Marquardt,
}

\usepackage{siunitx}

\usepackage[caption=false,font=footnotesize]{subfig}

\usepackage{ctable}

\title{\LARGE \bf
UVIO: An UWB-Aided Visual-Inertial Odometry Framework with Bias-Compensated Anchors Initialization
}

\author{
Giulio Delama$^{1}$, Farhad Shamsfakhr$^{2}$, Stephan Weiss$^{1}$, Daniele Fontanelli$^{2}$ and Alessandro Fornasier$^{1}$
\thanks{$^{1}$Giulio Delama, Stephan Weiss, and Alessandro Fornasier are with the Control of Networked Systems Group, University of Klagenfurt, Austria. {\tt \{giulio.delama,stephan.weiss, alessandro.fornasier\}@ieee.org} }
\thanks{$^{2}$Farhad Shamsfakhr and Daniele Fontanelli are with the Department of Industrial Engineering, University of Trento, Italy. {\tt \{farhad.shamsfakhr,daniele.fontanelli\}@unitn.it} }
\thanks{This work was supported by the Federal Ministry for Climate Action,
Environment, Energy, Mobility, Innovation and Technology (BMK) under the grant agreement 894790 (SALTO) and by the EU-H2020 project BUGWRIGHT2 (GA 871260).}
\thanks{\textbf{{Preprint version~\copyright IEEE, DOI: 10.1109/IROS55552.2023.10342012.}}}
}

\hypersetup{
  pdfinfo={
    Title={},
    Author={},
    Subject={},
    Keywords={}
  }
}

\begin{document}

\maketitle
\thispagestyle{empty}
\pagestyle{empty}


\begin{abstract}
  This paper introduces UVIO, a multi-sensor framework that leverages \ac{uwb} technology and \ac{vio} to provide robust and low-drift localization. In order to include range measurements in state estimation, the position of the \ac{uwb} anchors must be known. This study proposes a multi-step initialization procedure to map multiple unknown anchors by an \ac{uav}, in a fully autonomous fashion. To address the limitations of initializing \ac{uwb} anchors via a random trajectory, this paper uses the \ac{gdop} as a measure of optimality in anchor position estimation, to compute a set of optimal waypoints and synthesize a trajectory that minimizes the mapping uncertainty. After the initialization is complete, the range measurements from multiple anchors, including measurement biases, are tightly integrated into the \ac{vio} system. While in range of the initialized anchors, the \ac{vio} drift in position and heading is eliminated. The effectiveness of UVIO and our initialization procedure has been validated through a series of simulations and real-world experiments.
\end{abstract}
\section{Introduction}

Autonomous operation of mobile robots and \acp{uav} has seen substantial growth and advancement in recent years. With numerous applications ranging from industrial automation to the exploration of hazardous environments, the requirement for accurate and low-drift state estimation is crucial for ensuring the safe and efficient functioning of these systems. One of the biggest challenges in autonomous navigation is operating in \acs{gnss}-denied environments. In such scenarios, the need for robust and reliable alternative localization methods becomes increasingly important. \ac{vio} has emerged as a powerful approach for providing real-time state estimation in these scenarios. It uses a combination of visual information from one or multiple cameras and inertial information from \acp{imu} that incorporate accelerometers and gyroscopes. However, despite its successes, \ac{vio} suffers from accumulative drift over time due to the inherent limitations of visual-based localization methods. In addition \ac{vio} is prone to fail in poor visual conditions. To address this, recent research has explored the integration of \ac{uwb} range measurements into \ac{vio} systems, leveraging the global information provided by these sensors to improve localization performance. In this paper, we introduce UVIO, a multi-sensor framework built upon the well-known OpenVINS \cite{geneva2020openvins}, that combines the strengths of both sensor modalities to overcome the limitations of traditional \ac{vio} and achieve low-drift localization for mobile robots and \acp{uav}. Together with the automated detection and initialization of the \ac{uwb} anchors, UVIO provides a complete, easily deployable solution for real-time state estimation in \acs{gnss}-denied environments. The initialization procedure starts with a random flight with the objective of obtaining a first coarse estimation of the anchors' positions by solving a linear least squares problem followed by a nonlinear optimization on the same data. This initial estimate is then used in a refinement procedure that consists of computing and flying an optimal trajectory, collecting the data, and solving a nonlinear least squares problem to refine the initial solution. For this purpose, the \ac{gdop}, a metric tightly related to the \ac{fim}, is used as a measure of optimality for deriving a set of waypoints that minimizes the mapping uncertainty. The entire pipeline is carried out by the \ac{uav} itself (navigating initially only based on \ac{vio}), without the need for any human intervention. By incorporating the proposed initialization procedure, the UVIO framework is able to automatically estimate the position of the \ac{uwb} anchors, which can subsequently be included in the estimation process to improve the accuracy and reliability of the localization system. This allows the \ac{uav} to operate in a completely autonomous manner, without the need for human intervention, even in \acs{gnss}-denied environments. The efficacy of the proposed procedure is demonstrated through a series of simulations and real-world experiments, showcasing the robustness and reliability of the UVIO framework even in situations of poor visual conditions where pure \ac{vio} is prone to fail.

\section{Related Work}

Several studies have explored the use of \ac{uwb} measurements to enhance localization accuracy, which plays a key role in autonomous navigation. This section briefly reviews some of the most relevant works and discusses their significance in relation to our research, with a specific focus on studies that tackle the \ac{uwb} anchors' initialization problem. Early studies, such as~\cite{wang2017ultra},~\cite{grau2017multi}, and~\cite{tienmann2018enhanced}, employed \ac{uwb} measurements with prior knowledge of anchor positions, resulting in a drift-free estimation of the robot's pose, and showed the potential of this approach to improve current navigation and localization systems. These methods required an offline calibration of the \ac{uwb} anchors' initial position, reducing the autonomy of operation and making them unsuitable for large-scale and dynamic environments.

Recent studies, such as~\cite{yang2021uvip},~\cite{shin2021mirvio},~\cite{queralta2022vio},~\cite{zhan2022improving},~\cite{xu2022omni} and~\cite{lin2023gnssdenied}, aimed to improve localization accuracy and reduce drift by integrating \ac{uwb} range measurements and \ac{vio}. These studies rely on the assumption that the \ac{uwb} anchors' positions are known a priori.
They adopt graph-optimization methods for state estimation, as opposed to the filter-based method considered in this work. The choice of method depends on the specific requirements of the problem.  
Graph-based methods for state estimation can often provide more accurate results than filter-based methods, but at a higher computational cost. On the other hand, filter-based methods are typically less computationally intensive and can be more easily implemented in real-time applications where computational resources are limited.

The authors in~\cite{hausman2016self} proposed a multi-sensor fusion framework based on the \ac{ekf} to estimate anchors and \ac{uav} positions jointly. The initialization of the \ac{uwb} anchors is performed in a single step by solving a linear least-squares problem, which can lead to inaccurate or completely wrong initialization results.

\cite{shi2019anchor} presents an accurate and easy-to-use method for \ac{uwb} anchor self-localization by utilizing ranging measurements and readings from a low-cost \ac{imu}. The locations of the anchors are estimated by freely moving the tag, using a tightly-coupled \ac{eskf} to fuse the \ac{uwb} and inertial measurements. The anchors are first initialized via an Iterative Least-Squares without considering measurement biases.

In~\cite{ridolfi2021uwb}, a self-calibration algorithm for ultra-wideband positioning systems with active anchor nodes is shown. The algorithm uses iterative gradient descent and error detection from a convolutional neural network to estimate the positions of the fixed anchors with high accuracy, even in non-line-of-sight (NLOS) conditions, but requires the anchors to exchange range messages between each other.

The study in~\cite{gao2021low} addresses the drift in \ac{vio} for indoor localization by incorporating a single \ac{uwb} anchor into the localization process. The initialization of the anchor is performed by solving a nonlinear Least-Squares problem that requires an initial guess for the position of the anchor that must be provided and known a priori.

The studies of Nguyen et al. in~\cite{nguyen2020tightly} and~\cite{nguyen2021range} proposed two tightly-coupled odometry frameworks that combine monocular visual feature observations with distance measurements provided by a single \ac{uwb} anchor. The first framework uses a variant of the Levenberg-Marquardt non-linear optimization algorithm to simultaneously estimate the scale factor and the anchor position while the second framework leverages the \ac{uwb} measurements more effectively by addressing the time-offset of each range data and utilizing all available measurements. Later, in~\cite{nguyen2021viral} and~\cite{nguyen2022viral}, the authors propose a graph-based optimization approach for multi-sensor \ac{slam} for \acp{uav}. The system fuses data from various sensors, including \ac{imu}, cameras, lidars, and \ac{uwb} range measurements to estimate the state of the vehicle in real time and with long-term consistency.

A recent paper from Jia et al.~\cite{jia2022fejviro} proposes a Visual-Inertial-Ranging Odometry (VIRO) approach to reduce localization drift in \ac{vio} systems by incorporating \ac{uwb} ranging measurements from a minimum of 3 anchors. The method utilizes a long-short time-window structure to initialize the anchors' positions and covariance. The paper also analyzes the observability of the VIRO system with unknown \ac{uwb} anchor positions and shows that there are four unobservable directions in the ideal case, which are reduced by fusing the \ac{uwb} measurements with a \ac{msckf}.

All of the studies presented so far show the potential of including \ac{uwb} technology for localization and in particular the mitigation of the drift in \ac{vio} navigation systems. However, none of the above-cited studies considers the fact that the solution of anchor initialization, which in turn affects the accuracy of the state estimation, is highly dependent on various factors, including the geometry of the disposition of the anchors and the trajectory of the \ac{uav} or the mobile robot that collects the initialization data. In a recent study~\cite{blueml2021bias}, the authors applied the principles of \ac{fim} to determine the optimal points within a designated flight volume for collecting data for initializing the position of a single \ac{uwb} anchor together with its measurement biases. The method consists of a coarse triangulation using random vehicle positions, followed by fine triangulation using \ac{fim}-optimized points that provide maximal information for the single anchor.

Our approach extends and improves the ideas in~\cite{blueml2021bias} on information-driven \ac{uwb} anchor initialization and usage by developing a fully automated initialization approach for \emph{simultaneous} initialization and subsequent use of \emph{multiple initially unknown} \ac{uwb} anchors with measurement biases. We make use of the \ac{gdop} as a novel metric to guide the optimization of data collection, initialization, and uncertainty reduction for a scalable set of \ac{uwb} anchors. More precisely, our contributions are:
\begin{itemize}
    \item Fully automated procedure to detect, initialize and use initially unknown \ac{uwb} anchors in a vision-\ac{uwb} supported estimation approach. 
    \item Novel least squares formulation, dubbed the optimal double method (Sec.~\ref{ssec:LS}), increasing the matrix condition number for improved coarse anchors initialization. 
    \item Optimal waypoint generation in arbitrarily shaped volumes based on the \ac{gdop} metric to enable optimal data collection for precise and \emph{simultaneous} initialization of multiple \ac{uwb} anchors.
    \item Validation of the proposed approach in simulations and in real experiments with an autonomously flying \ac{uav} (and open sourcing of the code).
    \item Real-world demonstration of the resilience of our approach against camera dropouts.
\end{itemize}

\section{UWB-aided Visual Inertial Odometry Filter Design}


\subsection{UVIO filter state}
Our proposed EKF-based UVIO filter has the following structure.
Beside the classical \ac{msckf}-\ac{vio} \emph{core state}, $^I\textsc{x}$, including the orientation $\quat{I}{G}$, position $\Vector{G}{p}{I}$, and velocity $\Vector{G}{v}{I}$ of the \ac{imu} frame \frameofref{I} in the global frame \frameofref{G}, as well as the gyroscope and accelerometer bias $\Vector{}{b}{\omega}$, $\Vector{}{b}{a}$, 
the \emph{clones state} $^C\textsc{x}$, including the clones of the \ac{imu} pose at different past time steps, the \emph{features state} $^F\textsc{x}$, including the position of $N$ features in the global frame, and the \emph{calibration state}
$^W\textsc{x}$, including camera intrinsics and extrinsics for each camera present, 
the UVIO filter is augmented with the \emph{UWB state} $^U\textsc{x}$. The latter includes the UWB extrinsic parameter, that is the position of the UWB tag with respect to the IMU frame ${\Vector{I}{p}{U}}$, the position of $M$ anchors in the global frame ${\Vector{G}{p}{A_1}, \cdots, \Vector{G}{p}{A_M}}$, as well as the bias parameters  ${\beta_1, \gamma_1, \cdots, \beta_M, \gamma_M}$ for each anchor. The bias parameters include the so-called \emph{distance bias} and \emph{constant bias}, described in later sections.
Therefore, the UVIO state is denoted as
\begin{equation}
    \Vector{}{x}{} = \left(^I\textsc{x}, ^C\textsc{x}, ^F\textsc{x}, ^W\textsc{x}, ^U\textsc{x}\right).
\end{equation}

In the proposed filter, each UWB anchor can be treated as \emph{fixed}, i.e., their values are kept constant, or can be \emph{refined online} and hence added to the UWB state. With two anchors fixed, we can set a non-drifting global frame (in fact, a non-linear observability analysis shows that fixing one anchor in position and the heading toward another anchor would be sufficient; fixing two anchors in position is done to simplify the implementation in the results section).

\subsection{UVIO range update}
For each anchor $A$, a range measurement ${z_A = h(\Vector{}{x}{}) + n}$, is described by the following model:
\begin{equation}\label{eq:hx}
    \begin{split}
     h(\Vector{}{x}{}) &\coloneqq \beta d + \gamma = \beta \norm{\tagpos - \anchpos} + \gamma, \\
     &= \beta \norm{\Vector{G}{p}{I} + \Rot{I}{G}^T\Vector{I}{p}{U} - \anchpos} + \gamma,
     \end{split}
\end{equation}
where $\beta$ and $\gamma$, firstly introduced in~\cite{blueml2021bias}, are respectively a multiplicative scalar that acts as a distance-dependent bias, and a constant bias term.  

For each range measurement $z_A$ from anchor $A$, define 
\begin{equation*}
    \bm{\Gamma}(\hatVector{}{x}{}) = \hat{\beta}\, \frac{\left(\hatVector{G}{p}{I} + \hatRot{I}{G}^T\hatVector{I}{p}{U} - \hatanchpos\right)^T}{\norm{\hatVector{G}{p}{I} + \hatRot{I}{G}^T\hatVector{I}{p}{U} - \hatanchpos}},
\end{equation*}
then, the measurement Jacobian $\mathbf{H}_{z_A}$ is defined through the following derivatives evaluated at $\hatVector{}{x}{}$:
\begin{subequations}\label{eq:Hx}
    \begin{minipage}{0.675\linewidth}
        \begin{align*}
             \frac{\partial  h(\Vector{}{x}{})}{\partial \Rot{I}{G}} &= \bm{\Gamma}(\hatVector{}{x}{})\, \hatRot{I}{G}^T\, \skewsym{\hatVector{I}{p}{U}}, \\
             \frac{\partial  h(\Vector{}{x}{})}{\partial \Vector{I}{p}{U}} &= \bm{\Gamma}(\hatVector{}{x}{})\, \Rot{I}{G}^T,\\
             \frac{\partial  h(\Vector{}{x}{})}{\partial \beta} &= \norm{\hatVector{G}{p}{I} + \hatRot{I}{G}^T\hatVector{I}{p}{U} - \hatanchpos},
        \end{align*}
    \end{minipage}
    \begin{minipage}{0.3\linewidth}
        \begin{align*}
            \frac{\partial  h(\Vector{}{x}{})}{\partial \Vector{G}{p}{I}} &= \bm{\Gamma}(\hatVector{}{x}{}), \\
            \frac{\partial  h(\Vector{}{x}{})}{\partial \Vector{G}{p}{A}} &= -\bm{\Gamma}(\hatVector{}{x}{}), \\
            \frac{\partial  h(\Vector{}{x}{})}{\partial \gamma} &= 1.
        \end{align*}
    \end{minipage}
    \vspace{7.5pt}
\end{subequations}

\begin{figure}[!tbp]
\centering
\includegraphics[width=\linewidth]{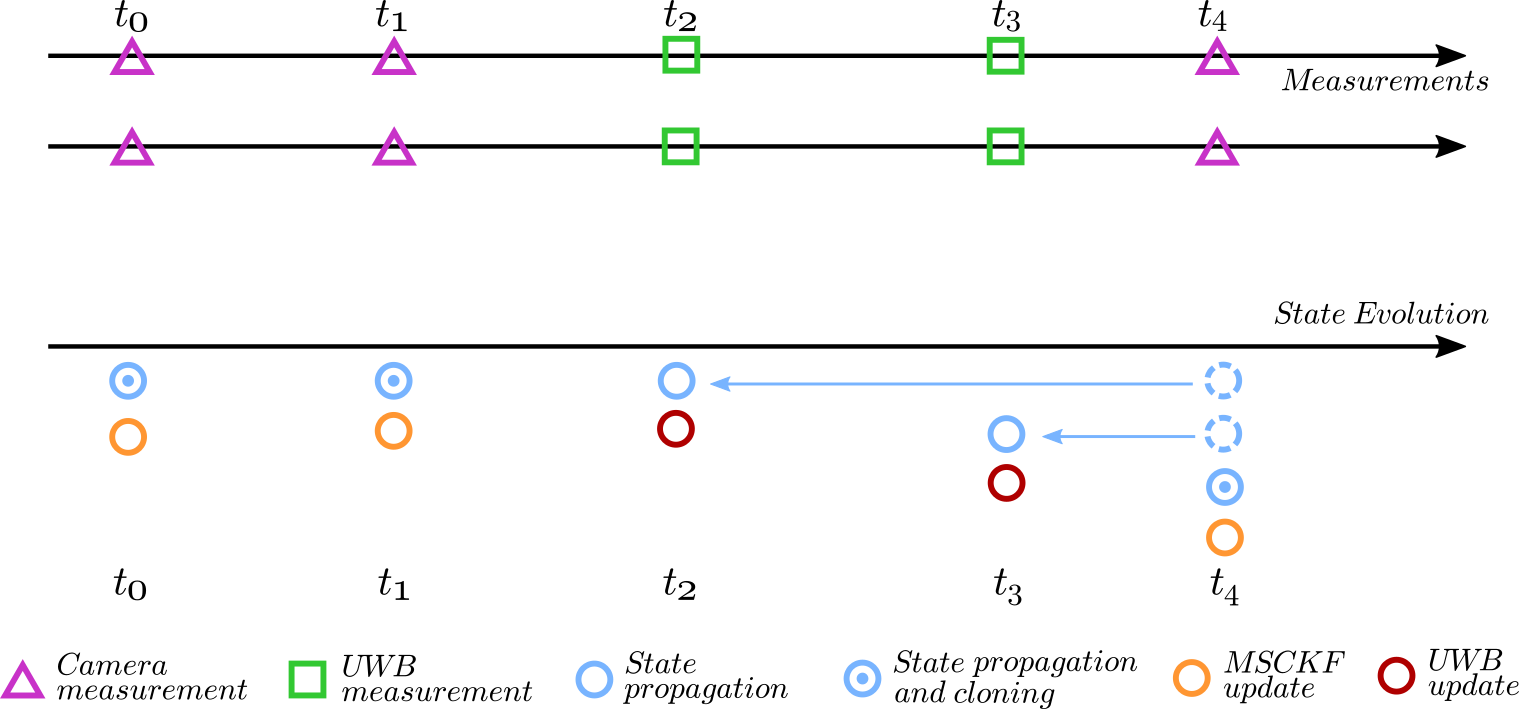}
\vspace{-5mm}
\caption{This figure shows an example of what we described as delayed update strategy. In this example, the actual state estimate is at the time the last camera measurement has been received, thus $t_1$. Two UWB measurements are received at time $t_2$ and $t_3$. At the next camera measurement, at time $t_4$, the state is propagated and updated through all the UWB measurements collected, and finally, at time $t_4$ a new clone is added and then the state is updated with the newest camera measurement.}
\label{fig:delayed_update}
\vspace{-5mm}
\end{figure}

Contrary to~\cite{jia2022fejviro}, we do not interpolate range measurements to align them with the camera measurements, but, to avoid introducing extra source of errors due to interpolation, we rather perform what we called \emph{Delayed Update}. As shown in \figref{delayed_update}, this update strategy consists of collecting range measurements in between the latest update at time $t_n$ and the next camera image at time $t_k$. Upon receiving a new camera image at time $t_k$, a series of chained propagation-update from time $t_n$ to time $t_k$ are performed.


\section{UWB Anchor Initialization}
\label{sec:init}
In order to be able to exploit UWB measurements in the UVIO filter an initial position of the UWB anchors needs to be provided. To minimize the setup time from the practitioner's perspective, the UVIO framework provides an anchors detection and self-initialization module. The anchors' self-initialization routine only assumes that the robot pose (through \ac{vio} in our case), as well as UWB range measurements, are available to the system. Firstly, data is collected from a short random or standard trajectory, which is used to obtain a rough initial solution by solving a linear least squares problem. Subsequently, a nonlinear optimization problem is built on the same data, and it is solved to refine the first coarse anchors positioning. The resulting anchor's position estimate is then used to compute a set of waypoints optimizing our proposed \ac{gdop} metric to maximize triangulation information for all involved \ac{uwb} anchors. The data from these optimal waypoints is then used in a nonlinear optimization problem to refine all \ac{uwb} anchors positions simultaneously.

\subsection{Coarse anchor positioning: Optimal Double Method}
\label{ssec:LS}
Obtaining a first rough estimation of the anchors' positions is fundamental for having an initial guess for the nonlinear optimization. In~\cite{hausman2016self} and~\cite{blueml2021bias}, the coarse anchors' initialization is modeled as a least square problem, which provides a fast initial guess of the anchors' positions. However, the least square formulation is in practice often ill-posed yielding poor solutions. In this respect,~\cite{hausman2016self} introduced a measurement inclusion test, where measurements are added to the least square problem only if the condition number of the problem is decreased by adding such a measurement. In this work, we reformulated the least square problem described in \cite{blueml2021bias} with what we called \emph{optimal double method}, aiming at increasing the condition number of the least square problem and minimizing the uncertainty of the solution.

The first choice we made to improve the conditioning of the least square formulation of the anchors' coarse initialization problem is to consider a known $\beta = 1$ in the measurement model in \equref{hx}, therefore, for a given range measurement $z$ from anchor $A$, we write
\begin{equation*}
     z = \norm{\tagpos - \anchpos} + \gamma + n \qquad n \sim \mathcal{N}\left(0, \sigma_n^2\right) .
\end{equation*}
Ignoring the measurement noise, we can square the previous equation and rewrite it as follows:
\begin{equation}
\label{eq:DistSquared}
\begin{aligned}
    z - \gamma &= \norm{\tagpos - \anchpos},\\
    \left(z - \gamma\right)^2 &= 
    \left(\tagpos - \anchpos\right)^T\left(\tagpos - \anchpos\right),\\
    z^2 &= \norm{\tagpos}^2 + \norm{\anchpos}^2 - 2\tagpos^T\anchpos + 2z\gamma - \gamma^2 .
\end{aligned}
\end{equation}
By applying the \emph{double method}~\cite{FontanelliSP21}, it is possible to derive a linear relation with respect to $\anchpos$ and $\gamma$. More precisely, consider $z_i$ and $z_j$ being two different (not necessarily consecutive) range measurements from the same anchor $A$, and ${{\tagpos}_i}$ and ${{\tagpos}_j}$ the \ac{uwb} tag positions at the time the measurements have been collected. Then by subtracting the two equations~\eqref{eq:DistSquared} we obtain
\begin{align*}
    z_i^2 - z_j^2 &= \norm{{\tagpos}_i}^2 - \norm{{\tagpos}_j}^2 - \\
    &- 2\left({\tagpos}_i - {\tagpos}_j\right)^T\anchpos + 2\left(z_i - z_j\right)\gamma,
\end{align*}
leading to the following least square formulation:
\begin{equation*}
\begin{bmatrix}
        \vdots\\
        \mathbf{A}_k\\
        \vdots
    \end{bmatrix} \Vector{}{x}{} = 
    \begin{bmatrix}
        \vdots\\
        b_k\\
        \vdots
    \end{bmatrix} \Rightarrow \mathbf{A} \Vector{}{x}{} = \Vector{}{b}{} ,
\end{equation*}
with
\begin{align*}
    \mathbf{A}_k &\coloneqq \begin{bmatrix}
        -\left({\tagpos}_i - {\tagpos}_j\right)^T & \left(z_i - z_j\right)
    \end{bmatrix} ,\\
    b_k &\coloneqq \frac{1}{2}\left(\left(z_i^2 - z_j^2\right) - \left(\norm{{\tagpos}_i}^2 - \norm{{\tagpos}_j}^2\right)\right) , \\
    \Vector{}{x}{} &\coloneqq \begin{bmatrix}
        \anchpos & \gamma
    \end{bmatrix}^T.
\end{align*}

In order to identify the pair $z_i$ and $z_j$, an optimal \emph{pivot} measurement to be subtracted from any other that minimizes the uncertainty of the solution should be determined. 
To this end, let ${n_k \sim \mathcal{N}\left(0, \sigma_n^2\right)}$, be the noise associated with the $k$-th range measurement from anchor $A$, and ${\varepsilon_k \sim \mathcal{N}\left(0, \bm{\Sigma}_{\varepsilon}\right)}$ be the noise associated with the $k$-th position of the UWB tag. In the analysis that follows, we consider only the vector $\Vector{}{b}{}$ being a stochastic entity while treating the matrix $\mathbf{A}$ as deterministic. Even though this is an approximation, it makes the analysis easier leading to a meaningful criterion for the choice of the optimal pivot.

Let us start by considering ${\mathbf{A}_k\Vector{}{x}{} = b_k = \bar{b}_k + \eta_k}$, with ${\eta_k \sim \mathcal{N}\left(0, \sigma_{\eta_k}^2\right)}$ and
\begin{equation*}
    \sigma_{\eta_k}^2 = \left(z_k^2 + z_p^2\right)\sigma_{n}^2 + {\tagpos}_k\bm{\Sigma}_{\varepsilon_k}{\tagpos}_k^T + {\tagpos}_p\bm{\Sigma}_{\varepsilon_p}{\tagpos}_p^T,
\end{equation*}
where $p$ is the index of the pivot.
Define ${\bm{\Sigma}_{\bm{\eta}} = \mbox{diag}\left[\sigma_{\eta_1}^2, \cdots, \sigma_{\eta_M}^2\right]}$ being the covariance of the vector ${\Vector{}{b}{}}$, then the information matrix of the least square solution is written
\begin{align*}
    &\bm{\Sigma}_{\Vector{}{x}{}}^{-1} = \bm{\mathcal{I}}_{\Vector{}{x}{}} =  \mathbf{A}^T\bm{\Sigma}_{\bm{\eta}}^{-1}\mathbf{A} = 
    \begin{bmatrix}
        \bm{\mathcal{I}}_{11} & \bm{\mathcal{I}}_{12}\\ \bm{\mathcal{I}}_{21} & \mathcal{I}_{22}
    \end{bmatrix},\\
    &\bm{\mathcal{I}}_{11} = \sum_{k = 1}^M \frac{\left({\tagpos}_k - {\tagpos}_p\right)\left({\tagpos}_k - {\tagpos}_p\right)^T}{\sigma_{\eta_k}^2},\\
    &\bm{\mathcal{I}}_{12} = -\sum_{k = 1}^M \left({\tagpos}_k - {\tagpos}_p\right)\frac{\left(z_k - z_p\right)}{\sigma_{\eta_k}^2},\\
    &\bm{\mathcal{I}}_{21} = -\sum_{k = 1}^M \frac{\left(z_k - z_p\right)}{\sigma_{\eta_k}^2}\left({\tagpos}_k - {\tagpos}_p\right)^T,\\
    &\mathcal{I}_{22} = \sum_{k = 1}^M \frac{\left(z_k - z_p\right)^2}{\sigma_{\eta_k}^2}.
\end{align*} 
Then, it turns out that the optimal pivot index $p$ can be chosen to fulfill the A-optimality criteria, that is such that the associated information matrix ${\bm{\mathcal{I}}_{\Vector{}{x}{}}}$ has maximum trace.


\subsection{Nonlinear Refinement}
\label{ssec:NLS}
The next step of our anchors' initialization routine is a nonlinear refinement of the first coarse solution based on the \ac{lm} algorithm, executed on the same data collected during the first random trajectory. In particular, we seek to minimize the following cost function:
\begin{equation}
    \min_{\Vector{}{x}{}} \frac{1}{2} \left\lVert\Vector{}{y}{}-\Vector{}{f}{}(\Vector{}{x}{})\right\rVert^{2},
\end{equation}
where $\boldsymbol{x}$ is the vector of unknown parameters to be estimated, $\Vector{}{y}{}$ is the vector of the measurement outputs, and $\Vector{}{f}{}(\Vector{}{x}{}) = \hat{\Vector{}{y}{}}$ is the vector of estimated outputs.
In contrast to the linear solution where only the constant bias $\gamma$ was estimated for each anchor, in the nonlinear optimization problem we include the distance-dependent bias $\beta$ for each anchor. The initial guess for the distance bias is set to $1$.

 In our problem formulation, the nonlinear relation between input and output is described by~\eqref{eq:hx}. For each anchor, the vector of parameters is defined to be $\Vector{}{x}{} = [\anchpos, \gamma, \beta]$ where $\anchpos$ is the anchor position in the global reference frame, $\gamma$ and $\beta$ respectively the constant and distant-dependent measurement bias.
The LM algorithm iteratively updates the values of $\boldsymbol{x}$ by linearizing the nonlinear function $\boldsymbol{f}(\boldsymbol{x})$ around the current estimate $\boldsymbol{x}_{k}$, and solving the resulting linear system $\Vector{}{J}{k}^{\mathrm{T}}(\Vector{}{y}{}-\Vector{}{f}{}(\Vector{}{x}{k})) \approx \Vector{}{0}{}$ where $\Vector{}{J}{k}$ is the Jacobian matrix at the current estimate. The algorithm adjusts the step size by introducing a scalar factor $\lambda_{k}$ that trades off between the steepest descent direction and the Gauss-Newton direction. The updated estimate is given by $\Vector{}{x}{k+1} = \Vector{}{x}{k} + \delta\boldsymbol{x}$ where
\begin{equation*}
    \delta\boldsymbol{x} = (\Vector{}{J}{k}^{\mathrm{T}}\Vector{}{J}{k} + \lambda_{k} \mbox{diag}(\Vector{}{J}{k}^{\mathrm{T}}\Vector{}{J}{k}))^{-1}\Vector{}{J}{k}^{\mathrm{T}}(\Vector{}{y}{}-\Vector{}{f}{}(\Vector{}{x}{k})).
\end{equation*}
The algorithm stops when either the change in the objective function $\left\lVert\Vector{}{y}{}-\Vector{}{f}{}(\Vector{}{x}{})\right\rVert^{2}$ is below a certain tolerance or the maximum number of iterations is reached. The estimated parameters covariance matrix of the final solution is:
\begin{equation*}
    \hat{\Vector{}{\Sigma}{}} = \mbox{MSE} (\Vector{}{J}{k}^{\mathrm{T}}\Vector{}{J}{k} + \lambda_{k} \mbox{diag}(\Vector{}{J}{k}^{\mathrm{T}}\Vector{}{J}{k}))^{-1}.
\end{equation*}

\subsection{Optimal Waypoints}
\label{ssec:OW}
\begin{figure}
  \centering
    \includegraphics[width=0.49\columnwidth]{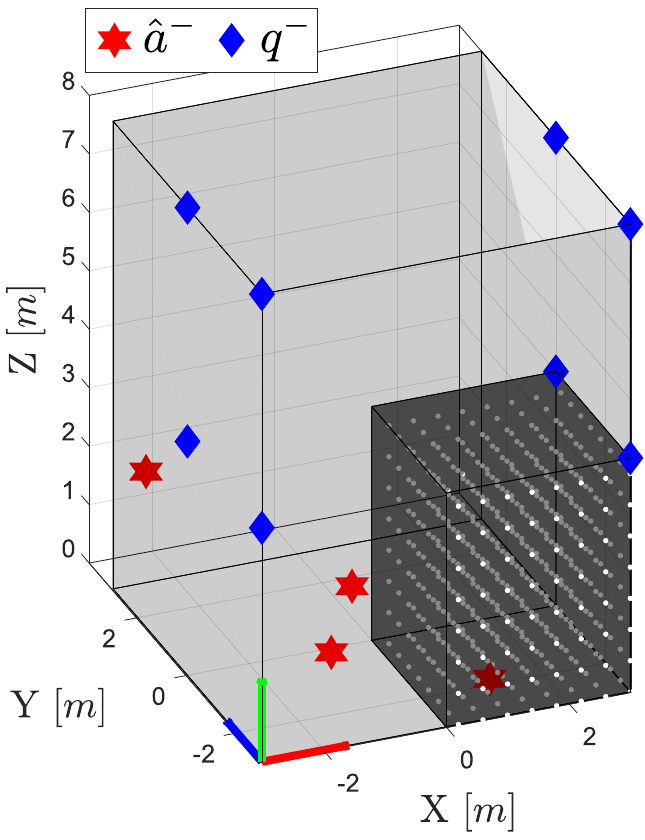}
\vspace{-4mm}
  \caption{The sample representation of the drone operational volume with 8 initial waypoints (i.e. $q^{-}$). The cube volume is divided into 8 smaller cubes (3D grids), each one indicating all the possible positions (denoted by small white dots) for the corresponding waypoint.}
  \label{fig:volume}
\vspace{-4mm}
\end{figure}

\begin{figure}[b!]
  \centering
    \includegraphics[width=0.7\columnwidth]{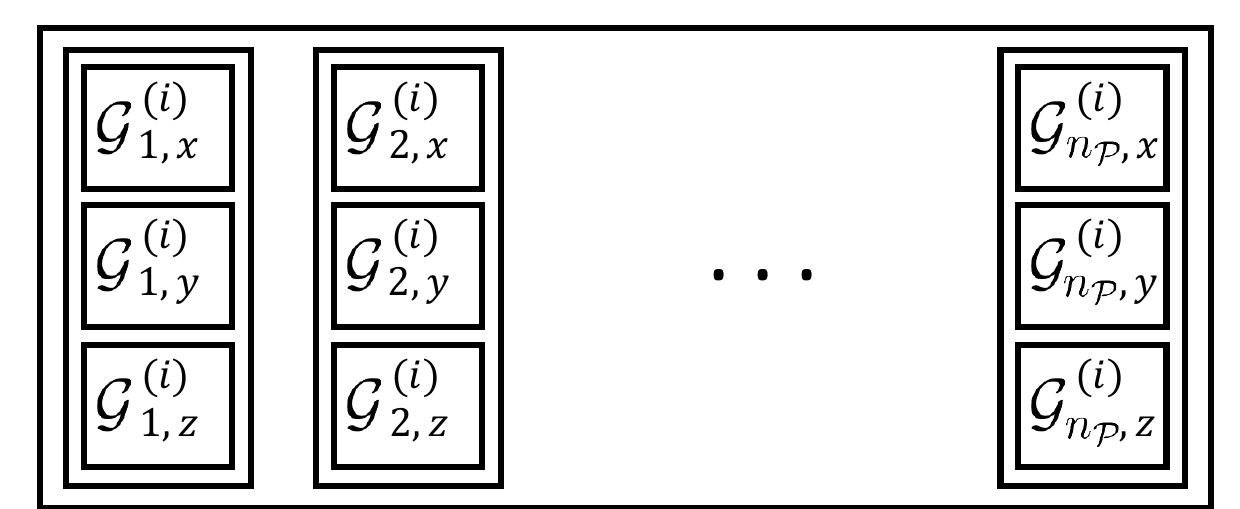}     
\vspace{-2mm}
    \caption{Schematic representation of a sample chromosome in the $i$th iteration of the Grid-Based Evolutionary Algorithm. $\mathcal{G}^{(i)}_{j,x},\,\mathcal{G}^{(i)}_{j,y},\,\mathcal{G}^{(i)}_{j,z}$ indicate the 3 indexes (xyz) of a sample grid
point for the $j$th waypoint.}
  \label{fig:chorom}
\end{figure}

Estimating the position of UWB anchors using the ranging measurements collected during a random trajectory has noticeable limitations despite the above introduced \emph{optimal double method}. The optimality of the trajectory with respect to information for anchor initialization can be evaluated based on different measures. 
In~\cite{blueml2021bias}, a first estimate of the anchor's position is refined by collecting data navigating through a set of waypoints computed via maximization of the Fisher information matrix. In this study, GDOP~\cite{FontanelliSP21} is employed as a measure of optimality for estimating the position of multiple anchors \emph{simultaneously}. A trajectory that minimizes the UWB mapping uncertainty is synthesized as follows.
Let $\mathbf{H}$ be the Jacobian in terms of the cosine directions of the nonlinear ranging measurements in~\equref{hx}. Considering one UWB anchor $A$ and $n_p$ waypoints we have
\begin{equation}
    \label{eq:MatH}
  \mathbf{H} = \begin{bmatrix}
   \cos\varphi_{1}\cos\gamma_{1}  & \cos\varphi_{1}\sin\gamma_{1}  &  \sin\varphi_{1} & 1\\
      \cos\varphi_{2}\cos\gamma_{2}  & \cos\varphi_{2}\sin\gamma_{2}  &  \sin\varphi_{2} & 1\\
      & \vdots  & \\
       \cos\varphi_{n_p}\cos\gamma_{n_p}  & \cos\varphi_{n_p}\sin\gamma_{n_p}  &  \sin\varphi_{n_p} & 1
  \end{bmatrix} ,
\end{equation}
where
\begin{align*}
    \gamma_{i} &= \arctan\left(\frac{{\anchpos}_y - q_y^{(i)}}{{\anchpos}_x - q_x^{(i)}}\right), \\
    \varphi_{i} &= \arctan\left(\frac{{\anchpos}_z - q_z^{(i)}}{\sqrt{({\anchpos}_x - q_x^{(i)})^2+({\anchpos}_y - q_y^{(i)})^2}}\right), 
\end{align*}
are, respectively, the Azimuth and Elevation angles of the $i$-th waypoint position $q^{(i)} = [q_x^{(i)}, q_y^{(i)}, q_z^{(i)}]$ with respect to the UWB anchor ${\anchpos}$. Now considering a minimum of four target positions, we can calculate the GDOP as follows:
\begin{equation}
    \label{eq:TrNLS}
\mbox{GDOP} = \sqrt{\mbox{Trace}(\mathbf{H}^T\mathbf{H})^{-1}} = \sqrt{\mbox{Trace}\bigg[\frac{\mbox{adj}(\mathbf{H}^T\mathbf{H})}{\det(\mathbf{H}^T\mathbf{H})}\bigg]},
\end{equation}
with
\begin{equation}
    \label{eq:DTNLS}
\det({\mathbf{H}^T \mathbf{H})} = (h_1 - h_2 + h_3 - h_4)^2,
\end{equation}
where
\[
  \begin{aligned}
    & h_t = -C_{\gamma_t}C_{\varphi_t}(C_{\varphi_k}S_{\gamma_k}\delta{s}_{ji}+C_{\varphi_j}S_{\gamma_j}\delta{s}_{ik}+C_{\varphi_i}S_{\gamma_i}\delta{s}_{kj}), \\
    & \delta{s}_{ij} = S_{\varphi_i}-S_{\varphi_j}, \quad C_{\gamma_i} = \cos(\gamma_i), \quad S_{\gamma_i} = \sin(\gamma_i), \\
    & C_{\varphi_i} = \cos(\varphi_i), \quad S_{\varphi_i} = \sin(\varphi_i), \\
  \end{aligned}
\]
and $\mbox{adj}(\mathbf{H}^T\mathbf{H})$ is readily defined by the co-factor of $\mathbf{H}^T\mathbf{H}$. 
Maximizing~\eqref{eq:DTNLS} is equivalent to minimizing the volume of the estimation uncertainty ellipsoid, which is 
a function of only the Azimuth and the Elevation angles of the waypoints with respect to the UWB anchor position, thus indicating those angles as the optimization parameters. Despite this remark, multiple anchors should be considered as well. 
In addition, operational search volumes for the waypoints can have any arbitrary shape (with UWB anchor locations potentially outside of the drone operational volume). As such, the waypoints determination turns into a challenging optimization problem, especially in unstructured environments. Moreover, instead of optimizing multiple trajectories separately for each anchor, as in~\cite{blueml2021bias}, we are interested in generating one optimal trajectory that minimizes the estimation accuracy for all the UWB anchors simultaneously. To this end, instead of angles, we consider the waypoint positions directly as the parameters of the following optimization problem:
\begin{equation}
\label{eq:minPos}
\min_{\boldsymbol{q}} \quad \overline{g}{(\boldsymbol{q})}, \qquad \textrm{s.t.} \quad \mathbf{I}(\boldsymbol{q})=1,
\end{equation}
where:
\[
\overline{g}{(\boldsymbol{q})} = \frac{1}{n_a} \sum_{i=1}^{n_a} \sqrt{\mbox{Trace}(\mathbf{H}_i^T\mathbf{H}_i)^{-1}},
\]
with Jacobian $\mathbf{H}_i$ given in~\eqref{eq:MatH} and computed 
for all the $n_a$ anchors, i.e., $\forall i \in [1,n_a]$, and $\mathbf{I}(\boldsymbol{q})$ is an indicator function to check the feasible volume of the waypoints (i.e., 
a 3D polygon~\cite{hormann2001point}) and being $\mathbf{I}(\boldsymbol{q})=1$ 
if a point is inside the volume, $\mathbf{I}(\boldsymbol{q})=0$ otherwise. To tackle this problem, we applied the grid-based evolutionary algorithm in~\cite{yang2013grid}. The algorithm starts with the drone operational volume considered as a cube with height $h$, width $w$ and length $\ell$, and $n_p$ initial waypoints (i.e. $q^{(1,-)},\,\dots,\,q^{(n_p,-)}$) (see~\figref{volume}). 
The cube volume is divided into $n_p = n_{h_p}\times n_{{\ell}_p}\times n_{w_p}$ smaller cubes, where $n_{h_p}$, $n_{{\ell}_p}$ and $n_{w_p}$ indicate the number of coordinate variations along $Z$ (height), $X$ (length) and $Y$ (width) axis. Each small cube is a 3D grid denoted by $\mathcal{G}_1\,\dots\,\mathcal{G}_{n_p}$, which defines all the possible positions for the corresponding waypoint. The evolutionary~\algoref{cap} has each chromosome defined as a $3\times n_p$ matrix whose genes indicate $3$ indexes along the three axes of a sample grid point for the corresponding waypoint. Each chromosome represents one particular configuration of the set of waypoints in the discrete space (see~\figref{chorom}). 
As shown in~\algoref{cap}, the Grid-Based Evolutionary Algorithm has as input the 3D environment configuration parameters and its center $c_v$, the initial estimates of the UWB anchors (i.e. $\hat{\anchpos}$) and the current position of the UWB tag (i.e. $\hat{\tagpos}_k$ at time $t_k$), and returns the optimal set of waypoints (i.e. $q^{\star}_1,\,\dots,\,q^{\star}_{n_p}$) that minimizes~\eqref{eq:TrNLS}. 
\begin{algorithm}[b!]
\small
\caption{Grid-Based Evolutionary Algorithm for estimating optimal set of way points}\label{alg:cap}
\begin{algorithmic}
\State $\textbf{Input :} (h,\,\ell,\,w,\,c_v,\hat{\anchpos},\hat{\tagpos}_k,n_p = n_{h_p}\times n_{{\ell}_p}\times n_{w_p})$
\State $\textbf{Output :} (q^{\star}_1,\,\dots,\,q^{\star}_{n_p})$
\State $\mathcal{G}_{[n_{h_p}\times n_{{\ell}_p}\times n_{w_p}]}=\textbf{GenerateCubeGrid}(h,\,\ell,\,w,\,c_v,\,n_p)$

\Require $n_{\mathcal{P}},\,\mathcal{P} =[\mathcal{C}_{1},\dots,\,\mathcal{C}_{n_{pop}}],\forall{\,i}\,\, \mathcal{C}_i\in{\mathcal{G}}$
\State $\mathcal{P} \gets Initialize(\mathcal{P})$
\While{$\text{termination criterion not fulfilled}$}
    \State $\mathcal{F}  \gets Fitness\_assignment(\mathcal{P})$  \Comment{based on \equref{minPos}}
    \State $\mathcal{P}' \gets Mating\_selection(\mathcal{P})$
    \State $\mathcal{P} \gets Variation(\mathcal{P}')$  \Comment{Cross-over, Mutation}
\EndWhile
\State $\boldsymbol{q^{\star}} \gets \underset{\mathcal{C}}{\text{min}} ~\mathcal{F}$
\end{algorithmic}
\end{algorithm}

\begin{figure}[t!]
  \centering
  \begin{tabular}{cc}
    \includegraphics[width=0.4\columnwidth]{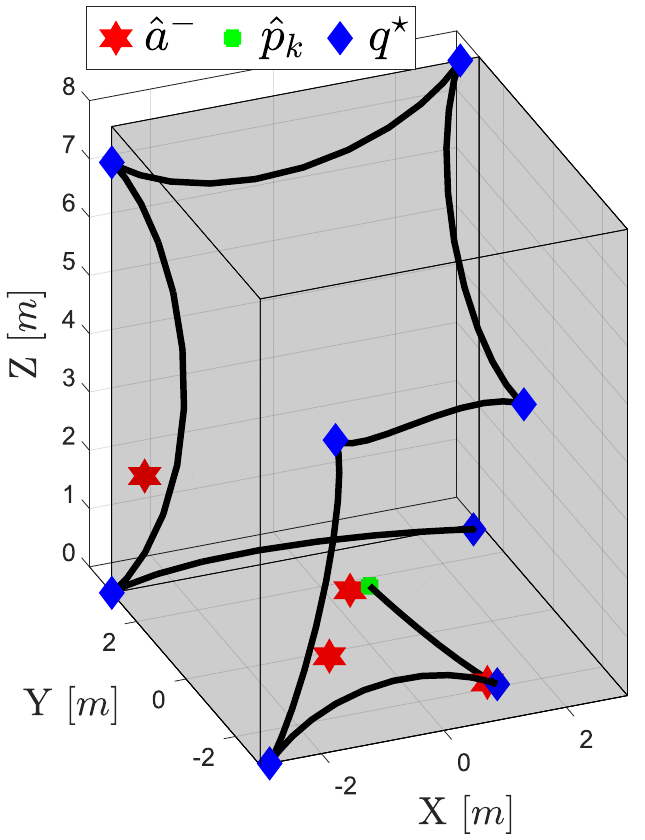} & \includegraphics[width=0.4\columnwidth]{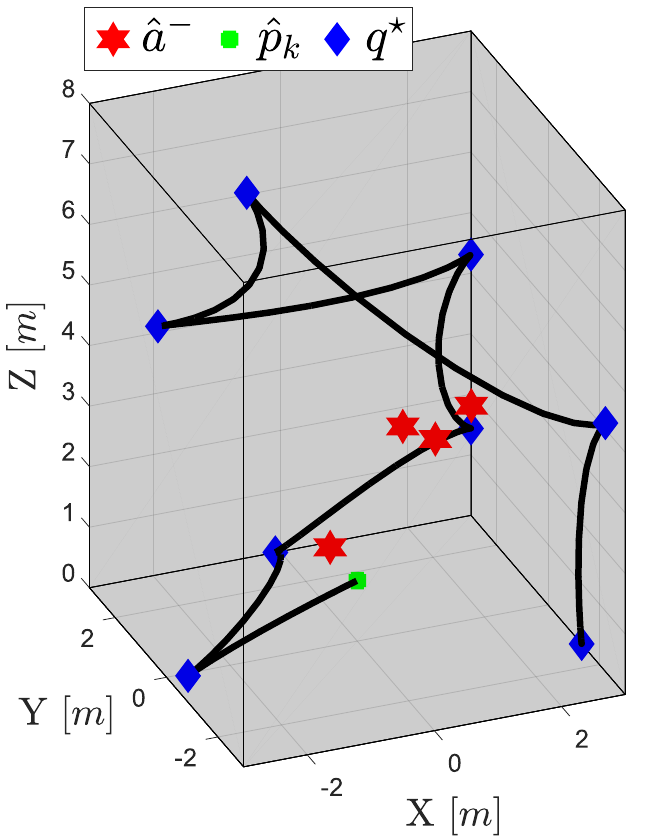} \\
    (a) & (b)\\
  \end{tabular}                                                    
  \caption{Two Optimal sets of waypoints estimated as the solutions of the Grid-Based evolutionary algorithm in two different random configurations of UWB anchors. The black curves show a minimum snap trajectory generated for the corresponding waypoints with zero velocity boundary conditions.}
  \label{fig:OptWP}
\end{figure}

\begin{figure}
  \centering
  \begin{tabular}{cc}
    \includegraphics[width=0.4\columnwidth]{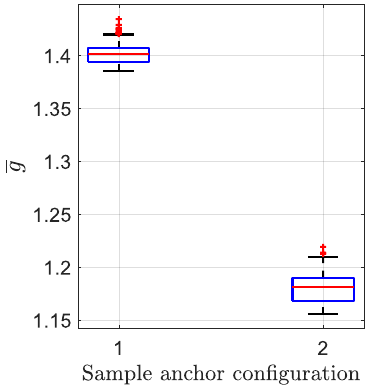} & \includegraphics[width=0.4\columnwidth]{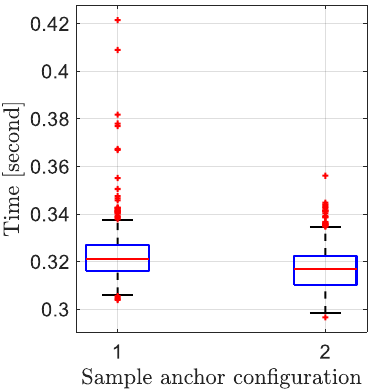} \\
    (a) & (b)\\
  \end{tabular}
\caption{Average GDOP and time distribution analysis for the two sample anchor configuration cases (1)~\figref{OptWP}(a) and (2)~\figref{OptWP}(b).}
  \label{fig:box_CostAndTime}
\vspace{-4mm}
\end{figure}

Here, we report the results of applying \algoref{cap} on two sample anchor configurations for two different random configurations of UWB anchors, reported in~\figref{OptWP}. The algorithm was executed with $2000$ maximum evolution of the genetic algorithm, tuned with a population size equal to $40$, crossover probability of $0.6$, mutation probability $0.3$ and elitism probability of $0.1$. The initial set of 8 waypoints was used which results in 8 sub-cubes as shown in~\figref{volume}, and the configuration space of each waypoint has $20^3$ different positions. The calculated optimal waypoints for the anchor configuration of~\figref{OptWP} (a) have the cost $\overline{g}(\boldsymbol{q}) = 1.4$, while for the corresponding anchor configuration of~\figref{OptWP} (b) $\overline{g}(\boldsymbol{q}) = 1.18$.
In order to have an idea of the repeatability of the solution provided by the Grid-Based Evolutionary Algorithm and of the corresponding execution time, we have carried out Monte Carlo (MC) simulations for the two sample anchor configurations shown in~\figref{OptWP} with $500$ trials. For the computation time analysis, we have used the same MC test and called the algorithm, which is written as a static C++ library and interfaced with Matlab Mex-Compiler, in each iteration of the MC trial. \figref{box_CostAndTime} shows the average GDOP and time distribution analysis for the two sample configuration cases (1)~\figref{OptWP}(a) and (2)~\figref{OptWP}(b).

After obtaining the set of optimal waypoints, the \ac{uav} can fly through them and collect data that will then be used again in a nonlinear refinement explained in Sec.~\ref{ssec:NLS} on the previously obtained solution. With this last step, we best leverage the information contained in the optimal waypoints previously calculated such that a precise initialization of all \ac{uwb} anchors in range can be performed \emph{simultaneously}.
\section{Experiments}
\subsection{Simulation}
\begin{figure}[t!]
    \centering
    \includegraphics[width=0.7\columnwidth]{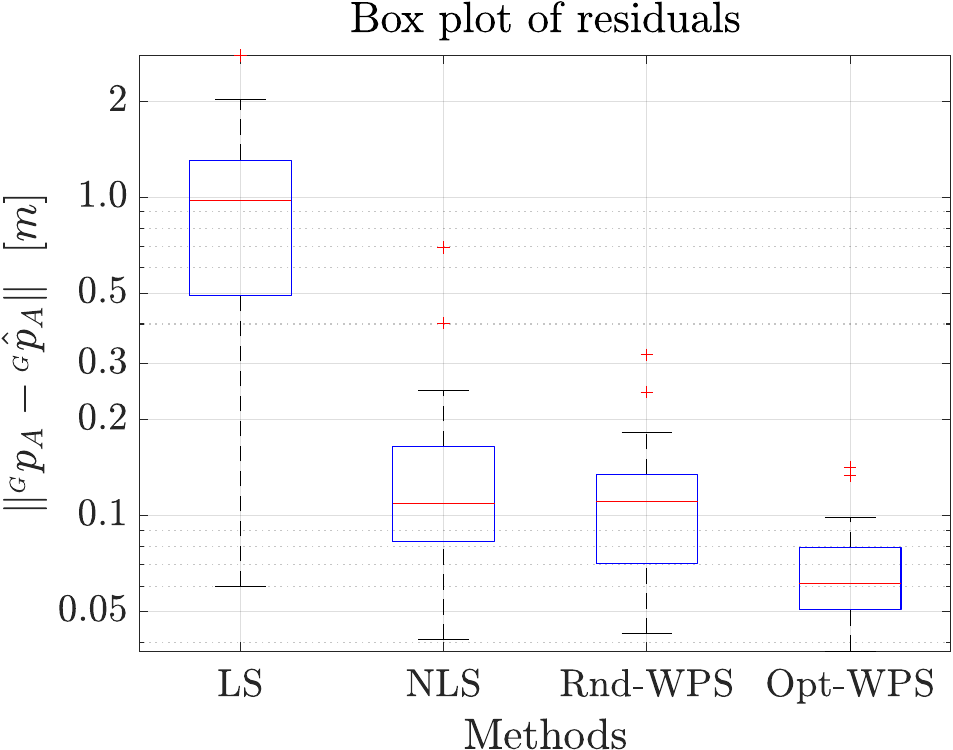}
\vspace{-2mm}
    \caption{Average position error for 10 different MC simulations of the full initialization procedure, with 100 realizations each, in a flight volume of $6\times6\times7~m^3$. Each simulation has a different initial trajectory and 4 or 5 randomly placed \ac{uwb} anchors with different biases. The measurement and position noise values are respectively $\sigma_y=0.3~m$ and $\sigma_p=0.03~m$.}
    \label{fig:simulation}
\vspace{-4mm}
\end{figure}

The initialization procedure was simulated and extensively tested in MATLAB, with a comprehensive exploration of multiple scenarios with different anchors configuration, \ac{uav} trajectories, parameter values, and noise levels. Particularly, in \figref{simulation} we compared the solutions of the initialization after each of the following steps: initial coarse least squares solution (LS), nonlinear least squares on the same initialization data (NLS), refinement of the nonlinear least squares via random waypoints (Rnd-WPS), and refinement via optimal waypoints (Opt-WPS). It is important to underline that the refinement via random and optimal waypoints is performed on "new" data with respect to the first two methods. For fairness of comparison, we considered the same amount of samples for all the methods. In all cases, the refinement procedure via optimal waypoints exhibited superior performance compared to other tested methods with notably improved minimization of both variance and mean error of the solutions. This outcome underscores the efficacy of incorporating an information-theoretic supported optimization-based approach for refining the initial solution of the anchor's mapping, increasing the reliability of the initialization process even in situations with challenging anchor configurations.

\subsection{Real-world experiments}
\begin{figure}[b]
    \centering
    \includegraphics[width=0.9\columnwidth]{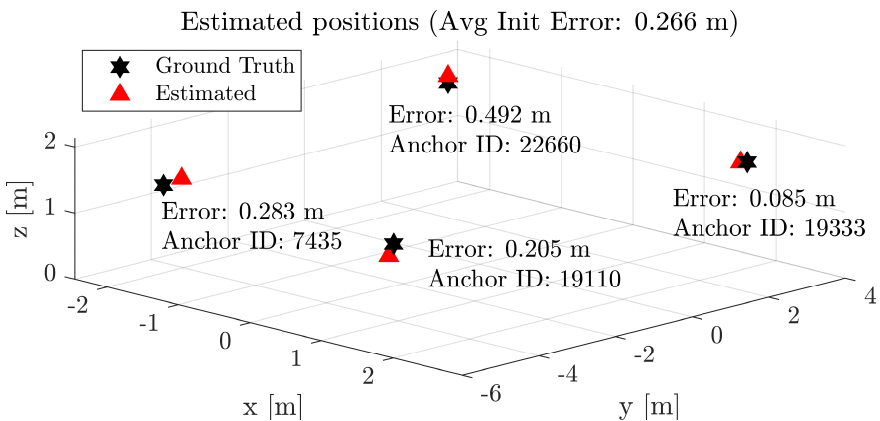}
\vspace{-3mm}
    \caption{Experiment 3: \ac{uwb} anchors initialization.}
    \label{fig:init}
\end{figure}
\begin{table}[b!]
\centering
\begin{tabular}{|c|c|c|c|c|c|}
\hline
Anc. ID & 19333 & 22660 & 7435 & 19110 & Avg.\\
\hline
Exp. 1 & $0.169~m$ & $0.470~m$ & $0.271~m$ & $0.193~m$ & $0.276~m$ \\
Exp. 2 & $0.153~m$ & $0.225~m$ & $0.150~m$ & $0.316~m$ & $0.211~m$ \\
Exp. 3 & $0.085~m$ & $0.492~m$ & $0.283~m$ & $0.205~m$ & $0.266~m$ \\
\hline
\end{tabular}
\caption{Initialization errors and averages.}
\label{tab:init_err}
\end{table}
\begin{figure}[t!]
  \centering
  \begin{tabular}{cc}
    OpenVINS: \\
    \includegraphics[width=0.98\columnwidth]{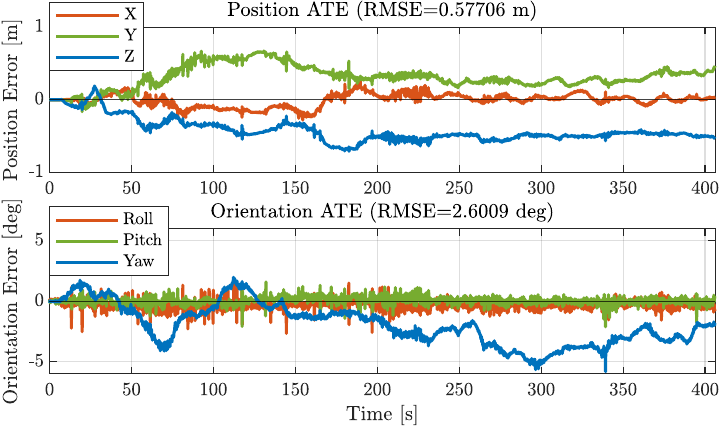} \\
    UVIO: \\
    \includegraphics[width=0.98\columnwidth]{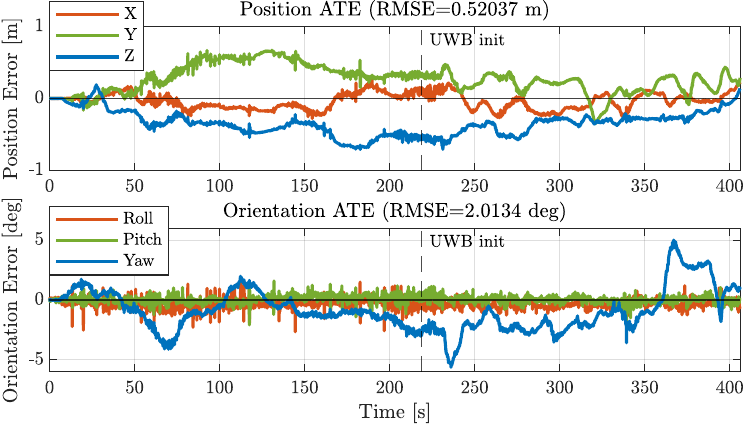} \\
  \end{tabular} 
\vspace{-4mm}
  \caption{Real-world experiment: comparison between OpenVINS and UVIO (real-time) position and orientation \ac{ate}. The \ac{rmse} of the whole trajectory is reduced by approximately $10\%$ in position and $20\%$ in orientation.}
  \label{fig:ATE}
\vspace{-4mm}
\end{figure}
In this section, we describe the real-world experiments conducted to validate the effectiveness of our proposed framework.  UVIO, coded in C++ extending OpenVINS, was run in real-time on the vehicle during the experiments. The experiments were carried out using a \ac{uav} equipped with an onboard flight computer (Raspberry Pi 4), an \ac{imu}, a camera (Matrix Vision BlueFOX), and a \ac{uwb} transceiver (Qorvo MDEK1001). The \ac{uav} was flown in a large indoor environment within a flight volume of $4\times6.5\times7~m^3$ equipped with an Optitrack motion capture system for recording the ground-truth poses. Four additional \ac{uwb} modules (Qorvo MDEK1001) were arbitrarily placed in the environment and their ground truth positions were captured. The system-autonomy framework introduced in \cite{flightstack} was used to enable the \ac{uav} to autonomously fly through waypoints inside the given flight volume. For the first set of experiments, we deployed the \ac{uav} at the center of the flight volume and executed the initialization procedure as described in Sec.~\ref{sec:init}. For the purpose of evaluating the accuracy of the proposed method, we used the ground-truth position of the \ac{uav} to perform the initialization. We performed three experiments with two configurations for the anchors and the corresponding position errors are shown in Tab.~\ref{tab:init_err}. The increase in error compared to the simulation experiments is explained by the non-ideality of the \ac{uwb} measurement model that assumes constant values for the biases during the initialization phase, while in reality interference, reflections, and other unwanted physical phenomena may change their values. Moreover, the results of the experiments show that the accuracy of the proposed method is influenced by the placement of the anchors with respect to the flight volume. In particular, anchors that are located outside of the volume and in a corner, result in higher initialization errors. On the other hand, anchors that are placed in favorable locations, such as near the center of a face of the flight volume, produce much lower initialization errors. A comparison with the results from \cite{blueml2021bias} suggests that the proposed approach is superior since it considers multiple anchors simultaneously and is capable of achieving a comparable or even lower initialization error for single anchors that are optimally placed, such as for anchor 19333 in \figref{init}.

\begin{figure}[t!]
    \centering
    \includegraphics[width=\columnwidth]{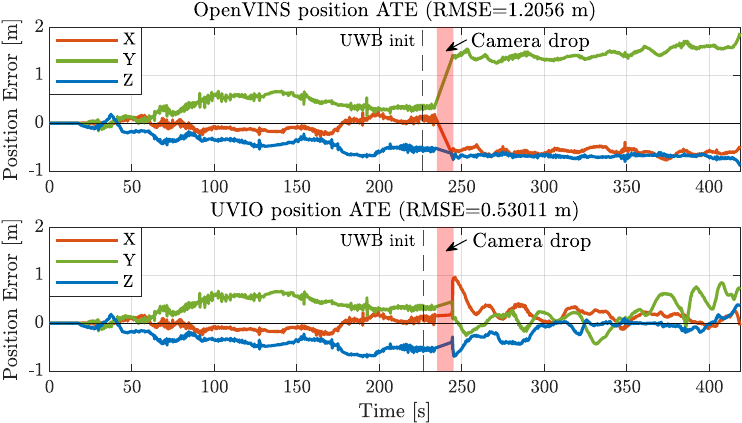}
\vspace{-7mm}
    \caption{UVIO and OpenVINS performance comparison with vision faults. With a full camera dropout of 10 seconds following the initialization of the \ac{uwb} anchors, after an initial transient caused by the \ac{vio} settling, UVIO is capable of fully recovering from the dropout.}
    \label{fig:ATEdrop}
\vspace{-4mm}
\end{figure}

A second set of experiments was carried out to evaluate the real-time performance of UVIO. This time, the initialization was performed considering the poses estimated by the \ac{vio} framework and stopped with the \ac{uav} hovering in place. This was followed by a manual flight of roughly the same duration, allowing for more agile maneuvering compared to the automatic flight through waypoints. For the evaluation of the trajectories, 
we chose the \ac{ate} as a metric~\cite{zhang2018tutorial} for the assessment of the UVIO performance. In all the experiments, we noticed an overall improvement in localization performance using UVIO compared to a \ac{vio}-only. This was reflected in a decrease of the \ac{ate} \ac{rmse} (\figref{ATE}). To further test and validate the reliability of our framework when compared to pure \ac{vio}, we simulated severe drops in framerate and full dropouts of the camera (\figref{ATEdrop}). The results of these experiments proved UVIO to be effective also in situations of poor visual conditions or camera faults, where pure \ac{vio} would fail.

\section{Conclusion}
In conclusion, this paper proposes UVIO, a multi-sensor framework that combines \ac{uwb} technology and \ac{vio} to provide reliable and low-drift localization for mobile robots and \acp{uav} operating in GNSS-denied environments. The proposed information-driven initialization procedure enables the automatic detection and mapping of multiple unknown anchors simultaneously, even in situations with challenging anchor configurations. The effectiveness of UVIO is validated through simulations and real-world experiments, by running it in real-time on a \ac{uav}, proving it to be a complete and easily-deployable solution for real-time state estimation in challenging scenarios where pure \ac{vio} is prone to fail. Future work could focus on further improving the initialization procedure by including a robust outlier detection for incoming \ac{uwb} ranges and expanding the applicability of the proposed approach to other robotic systems.

\bibliographystyle{IEEEtran}
\bibliography{bibliography/Reference.bib}

\end{document}